\documentclass[twocolumns]{M-PSI}
\usepackage[utf8]{inputenc}
\usepackage[T1]{fontenc}

\usepackage{graphicx}

\title{Calibration-Free 3D Multi-Camera People Tracking for Indoor Environment}

\author{
\AuthorEntry{Ponleur Veng}{,}{1,2}{}{0009-0006-1845-4969}
\AuthorEntry{Dominique Vaufreydaz}{,}{2}{https://research.vaufreydaz.org/}{0000-0002-8825-0973}
\AuthorEntry{Phutphalla Kong}{}{1}{}{0000-0003-1659-6745}\\
\vspace{0.5em}
\textnormal{\normalsize{
{$^1$ Cambodia Academy of Digital Technology}\\ %
{$^2$ Univ. Grenoble Alpes, CNRS, Grenoble INP, LIG, 38000 Grenoble, France}\\
}}
\vspace{0.5cm}
} %

\begin{document}
\begin{abstract}
Multi-Camera People Tracking (MCPT) traditionally relies on precise intrinsic and extrinsic camera calibration to project 2D detections into a unified 3D world coordinate system. However, manual calibration constitutes a major bottleneck in large-scale dataset generation from unconstrained video archives. This work proposes a unified calibration-free 3D MCPT framework that infers geometric structure directly from visual data using deep foundation models. The system integrates anchor-free detection (YOLOX), robust tracking (BoT-SORT), omni-scale appearance embedding (OsNet), pose estimation (HRNet via MMPose), and transformer-based geometric reconstruction using the Visual Geometry Grounded Transformer (VGGT). A pose-guided 3D lifting strategy projects head keypoints onto a reconstructed manifold, eliminating dependence on ground-plane homography. Global identity association is formulated as hierarchical agglomerative clustering under a joint appearance–geometry cost with strict velocity gating. Evaluation on the AI City Challenge 2024 demonstrates a HOTA score of 53.13\% without access to ground-truth calibration matrices, establishing a strong baseline for purely vision-based 3D tracking.

\end{abstract}
\keywords{Multi-camera tracking \and 3D geometry \and Calibration-free reconstruction \and Foundation models \and Pose-guided lifting \and HOTA}

\section{Introduction}
Understanding human motion in three-dimensional space is fundamental to modern computer vision systems deployed in real-world environments. Multi-Camera People Tracking (MCPT) extends single-view tracking by reconstructing consistent trajectories across spatially distributed cameras, enabling long-term identity preservation and scene-level behavioral analysis \cite{Ciaparrone_2020_survey}. Such capability is essential for intelligent surveillance, activity analytics, and smart learning environments.

Despite substantial progress in detection, tracking, and re-identification \cite{Bewley_2016,wojke2017simpleonlinerealtimetracking}, practical deployment of MCPT systems remains constrained by a critical bottleneck: camera calibration. Most existing 3D tracking frameworks assume accurate intrinsic and extrinsic parameters to project 2D detections into a unified world coordinate system \cite{Hartley_Zisserman_2004}. However, acquiring these calibration parameters in real environments typically requires manual procedures such as checkerboard-based calibration or multi-view geometric alignment \cite{888718,7780814}. This requirement severely limits the scalability of 3D tracking systems and restricts their applicability to controlled installations.

Indoor environments such as classrooms further amplify these challenges. Individuals are frequently seated, partially occluded by desks, and densely arranged in overlapping fields of view. Traditional ground-plane homography assumptions fail in such settings, particularly when lower-body keypoints are invisible. As a result, both geometric consistency and identity association degrade significantly under occlusion, as widely observed in crowded multi-camera tracking benchmarks \cite{huang2023enhancingmulticamerapeopletracking,10677892_sjtu_posetrack}.

Recent advances in large-scale geometric foundation models suggest an alternative direction. Instead of explicitly calibrating cameras, it is now possible to infer relative pose and scene structure directly from image data \cite{wang2024dust3rgeometric3dvision,wang2025vggtvisualgeometrygrounded}. These models learn dense 3D representations and camera relationships in a data-driven manner, reducing reliance on explicit calibration pipelines. This shift raises an important question: can multi-camera 3D tracking be achieved without any manual calibration, while remaining competitive with fully calibrated systems?

In this work, we explore this question by proposing a unified calibration-free 3D MCPT framework. Our approach leverages transformer-based visual geometry reconstruction to recover relative camera poses and dense scene structure directly from video frames \cite{wang2025vggtvisualgeometrygrounded}. To address common occlusion issues in indoor environments, we introduce a pose-guided 3D lifting strategy that projects anatomically stable head keypoints onto the reconstructed 3D manifold, eliminating reliance on foot-point or ground-plane assumptions. Global identity association is then formulated as a joint appearance–geometry clustering problem under strict physical velocity constraints.

We evaluate our framework on the AI City Challenge 2024 benchmark \cite{wang20248thaicitychallenge}, which provides large-scale, densely annotated 3D ground truth and evaluates performance using the Higher Order Tracking Accuracy (HOTA) metric \cite{Luiten_2020_hota}. Without access to ground-truth calibration matrices, the proposed method achieves a HOTA score of 53.13\%, outperforming several calibrated baselines. These results demonstrate that calibration-free geometry inference can serve as a viable foundation for scalable 3D multi-camera tracking.

Our contributions are summarized as follows:
\begin{enumerate}
    \item We propose a fully calibration-free 3D multi-camera tracking framework that infers scene geometry directly from visual data.
    \item We introduce a pose-guided 3D lifting strategy that improves robustness under severe lower-body occlusion.
    \item We demonstrate that foundation model-based geometric reconstruction enables competitive performance without manual calibration.
\end{enumerate}
We believe this work provides an important step toward scalable, geometry-aware tracking systems deployable in unconstrained real-world environments.

\section{Related Works}
\subsection{Single-Camera Perception}
The performance of any tracking-by-detection system is strictly upper-bounded by the quality of the initial pedestrian localizations \cite{Ciaparrone_2020_survey}. Early deep learning approaches, such as Faster R-CNN \cite{ren2016fasterrcnnrealtimeobject} and YOLOv3 \cite{redmon2018yolov3incrementalimprovement}, relied on anchor-based mechanisms. These models struggle with intra-class occlusion in crowded environments. Consequently, research has shifted toward anchor-free detectors. Methods like FCOS \cite{tian2019fcosfullyconvolutionalonestage} and CenterNet \cite{zhou2019objectspoints} formulated detection as a pixel-wise prediction task. YOLOX \cite{ge2021yoloxexceedingyoloseries} represents a mature realization of the anchor-free paradigm, offering improved convergence stability and detection quality in crowded scenarios compared to earlier YOLO variants such as its predecessors like YOLOv5 \cite{khanam2024yolov5deeplookinternal}.

For generating tracklets, SORT-based algorithms dominate the space. The seminal SORT algorithm \cite{Bewley_2016} utilized the Kalman Filter for motion prediction, while DeepSORT \cite{wojke2017simpleonlinerealtimetracking} extended this by incorporating a deep appearance descriptor. BoT-SORT \cite{aharon2022botsortrobustassociationsmultipedestrian} addresses failures in dynamic indoor scenes by integrating Camera Motion Compensation (CMC) and prioritizing accurate geometric association.

\subsection{Multi-Camera Association and Re-Identification}
Associating tracklets across disjoint camera views requires robust Person Re-Identification (ReID). Standard Convolutional Neural Networks (CNNs) like ResNet-50 \cite{he2015deepresiduallearningimage} often struggle to capture fine-grained features across varying scales. To mitigate this, OsNet \cite{zhou2019omniscalefeaturelearningperson} was introduced to dynamically fuse features from multiple scales. However, appearance embeddings are often insufficient alone to resolve complex multi-camera associations caused by heavy occlusion, necessitating Spatio-Temporal Constraints as seen in the winning solutions of the AI City Challenge 2023 \cite{huang2023enhancingmulticamerapeopletracking} and 2024 \cite{10677892_sjtu_posetrack}.

\subsection{Calibration-Free 3D Geometry}
To implement spatio-temporal constraints, systems must translate 2D coordinates into a unified 3D world, mathematically requiring camera calibration \cite{Hartley_Zisserman_2004}. Traditionally, solving for camera matrices involves rigorous manual calibration procedures using physical checkerboards \cite{888718} or traditional Structure-from-Motion (SfM) pipelines like COLMAP \cite{7780814}, which limit scalable deployment.

Recent advances in Deep Geometric Learning have introduced calibration-free 3D reconstruction. Foundation models such as DUSt3R \cite{wang2024dust3rgeometric3dvision} and the VGGT \cite{wang2025vggtvisualgeometrygrounded} represent the cutting edge of this domain. These models predict dense 3D point maps and spatial priors directly from sparse, wide-baseline views without explicit calibration steps.

\subsection{Benchmarks for Multi-Camera Tracking}
The validation of MCPT systems has historically relied on datasets evaluating 2D bounding box overlap, such as DukeMTMC \cite{ristani2016performancemeasuresdataset}. To properly evaluate the geometric accuracy of modern 3D tracking systems, benchmarks like the AI City Challenge 2024 \cite{wang20248thaicitychallenge} provide dense 3D ground truth and utilize the 3D Higher Order Tracking Accuracy (HOTA) metric \cite{Luiten_2020_hota}.

\section{Methodology}
\begin{figure*}
\centering
\includegraphics[width=0.85\textwidth]{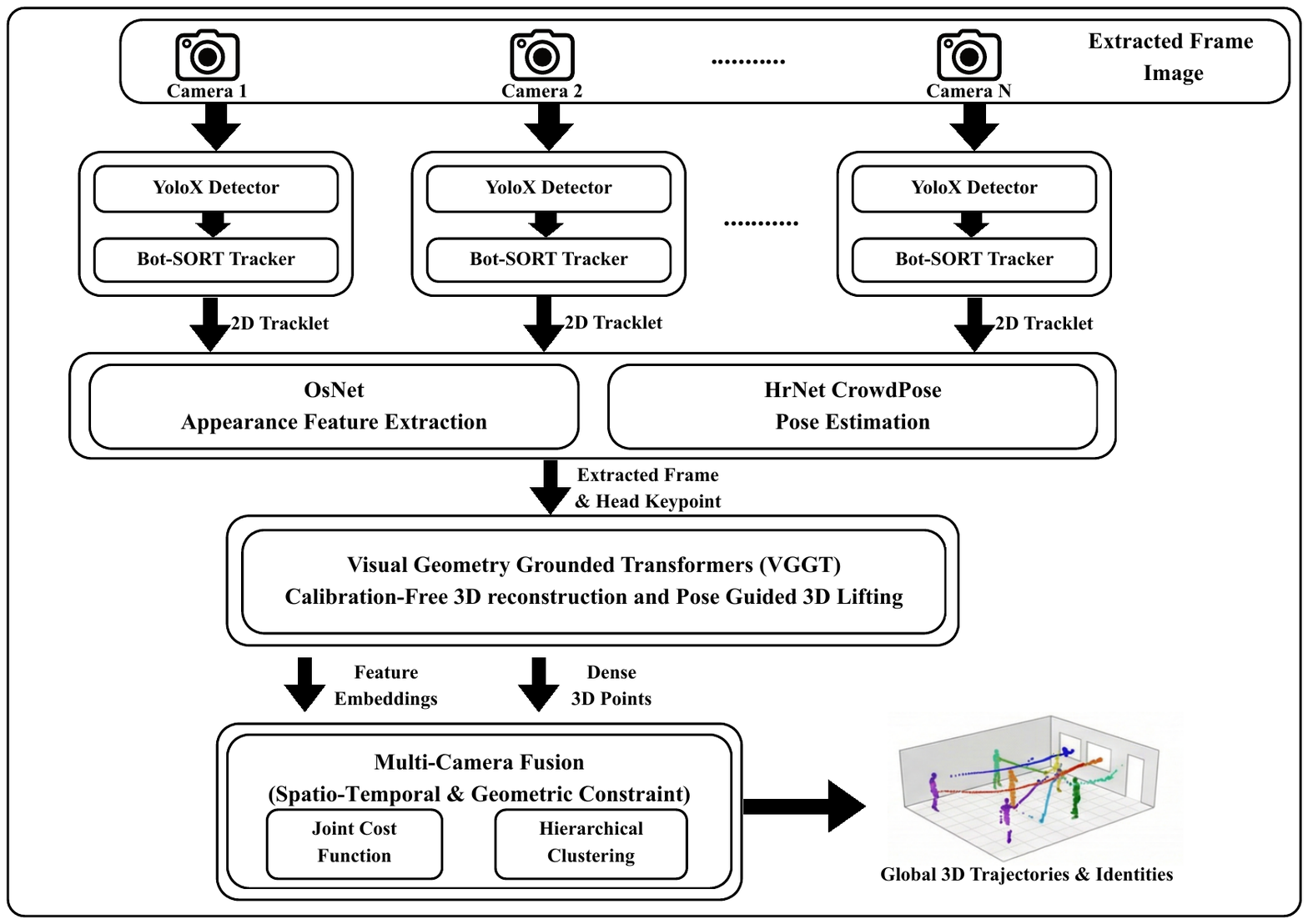}
\caption{System overview.} \label{system_overview}
\end{figure*}

The proposed pipeline transforms uncalibrated 2D video feeds into a unified 3D trajectory space through four sequential modules as indicated in Fig \ref{system_overview}.

\subsection{Robust 2D Tracking and Feature Extraction}
The input consists of synchronized video streams from $N$ static cameras. We utilize the YOLOX anchor-free detector to output pedestrian localizations:
\begin{equation}
    d_{i}=(x,y,w,h,s)
\end{equation}
where $(x,y)$ represents the top-left corner and $s$ is the confidence score. To generate coherent tracklets, BoT-SORT is employed. The Kalman Filter state vector incorporates bounding box differentials to predict positions during brief occlusions:
\begin{equation}
    x=[x_{c},y_{c},w,h,\dot{x}_{c},\dot{y}_{c},\dot{w},\dot{h}]
\end{equation}
For ReID, we extract visual embeddings using OsNet. To represent a tracklet robustly, we compute the averaged, unit-normalized feature vector across all frames in the tracklet:
\begin{equation}
    F_{tracklet}=\frac{\sum_{k=1}^{L}f_{app}^{(k)}}{||\sum_{k=1}^{L}f_{app}^{(k)}||}
\end{equation}

\subsection{Pose-Guided 3D Geometry (VGGT)}
Traditional methods rely on bounding box foot points, which fail during lower-body occlusions. We deploy an HRNet-w48 \cite{sun2019deep_hrnet} backbone trained on the CrowdPose \cite{Li_2019_CVPR_crowdpose} dataset using the MMPose framework \cite{mmpose2020} to extract the anatomical Head Keypoint. This is preferred over standard conventions like COCO \cite{lin2014microsoft_coco}, as it is explicitly designed to address the challenge of overlapping individuals.

Simultaneously, the VGGT-1B model \cite{wang2025vggtvisualgeometrygrounded} processes pairs of uncalibrated video frames to predict relative camera poses (rotation $R$ and translation $T$) and a dense 3D point cloud.
\begin{equation}
    P_{3D}^{head}=VGGT_{query}(u_{head},v_{head})
\end{equation}
This allows the system to recover the 3D head position directly from the reconstructed scene geometry, avoiding the common but fragile assumption of a flat ground plane.

\subsection{Multi-Camera Fusion and Optimization}
We associate local tracklets $T_{local}$ into global trajectories $T_{global}$ using a Joint Cost Function combining appearance and geometric distances:
\begin{equation}
    Cost(i, j) = W_{app} D_{app}(i,j) + W_{geo} D_{geo}(i,j)
\end{equation}
where $D_{app}$ is the Cosine Distance of OsNet vectors, and $D_{geo}$ is the normalized Euclidean distance between 3D head points:
\begin{equation}
    D_{geo}=min(\frac{||P_{3D}^{i}-P_{3D}^{j}||_{2}}{\lambda_{space}},1.0)
\end{equation}

To prevent physically impossible matches, we calculate the required transition speed $V_{trans}$ between tracklets. If $V_{trans} > 4.0m/s$, a heavy penalty cost of 10.0 is applied. Final identities are resolved via Hierarchical Agglomerative Clustering (HAC) using Complete Linkage, with the clustering dendrogram thresholded at $\tau_{cluster}=0.5$.

Intuitively, the appearance term encourages visual identity consistency, while the geometric term enforces physical plausibility in 3D space. The normalization factor $\lambda_{space}$ ensures that spatial distances are comparable in scale to appearance similarity scores.

\section{Experiments and Results}
\subsection{Dataset and Evaluation Metrics}
\begin{table}[t]
\caption{Statistics of the AI City Challenge 2024 (Track 1) dataset.}
\label{tab:dataset_stats}
\centering
\begin{tabular}{l c}
\hline
\textbf{Attribute} & \textbf{Value} \\
\hline
Total scenes & 90 \\
Total cameras & 953 \\
Total identities & 2,491 \\
Total bounding boxes & $> 100$ million \\
Video resolution & $1920 \times 1080$ \\
Frame rate & 30 FPS \\
Total duration & 212 hours \\
\hline
\end{tabular}
\end{table}

We utilize the Higher Order Tracking Accuracy (HOTA) metric, evaluated in physical world coordinates:

\begin{equation}
    HOTA_{\alpha}=\sqrt{DetA_{\alpha}\cdot AssA_{\alpha}}
\end{equation}

HOTA\cite{Luiten_2020_hota} jointly evaluates detection and association performance by combining spatial localization and identity consistency into a single geometric mean. The Spatial Detection Accuracy (DetA) component measures how precisely predicted trajectories align with ground-truth positions within a predefined distance threshold $\tau_{dist}$, effectively capturing spatial correctness in 3D space. The Global Association Accuracy (AssA) term evaluates the temporal stability of identities across frames and camera views, penalizing identity switches and fragmentation. Finally, the Localization Accuracy (LocA) metric assesses the fine-grained spatial alignment of correctly matched true positives with their corresponding ground-truth annotations. Together, these components provide a comprehensive evaluation of both geometric accuracy and identity consistency in multi-camera tracking systems.

\subsection{Ablation Study}
We conducted an ablation study to analyze the relative contribution of appearance ($W_{app}$) and geometric ($W_{geo}$) terms within the joint association cost function, as presented in Table \ref{tab:ablation_weights}.

When the model relies exclusively on appearance information ($W_{app}=1.0$, $W_{geo}=0.0$), it achieves a high association accuracy of 93.12\%, indicating that OsNet embeddings are highly discriminative for identity matching. However, the detection accuracy drops significantly to 30.55\%, demonstrating that appearance features alone fail to enforce sufficient spatial consistency under severe occlusion and crowded indoor conditions.

Conversely, when the model depends solely on geometric constraints ($W_{app}=0.0$, $W_{geo}=1.0$), the spatial detection accuracy improves to 33.71\%, reflecting the stabilizing effect of 3D proximity constraints. Nevertheless, association accuracy decreases substantially to 79.53\%, confirming that geometry alone lacks the discriminative power required to reliably distinguish visually similar individuals across cameras.

The optimal configuration is achieved through balanced fusion ($W_{app}=0.5$, $W_{geo}=0.5$), yielding the highest overall HOTA score of 53.13\%. This result demonstrates that appearance and geometry provide complementary cues, and their integration enables robust cross-view identity association while maintaining spatial coherence. Notably, the Localization Accuracy (LocA) remains constant at 62.18\% across all configurations, indicating that localization precision is primarily determined by the underlying YOLOX detector rather than the global association mechanism.

\begin{table}[t]
\caption{Impact of appearance ($w_{app}$) and geometry ($w_{geo}$) weights on tracking performance (validation set).}
\label{tab:ablation_weights}
\centering
\begin{tabular}{c c c c c c}
\hline
\multicolumn{2}{c}{\textbf{Fusion Weights}} & \multicolumn{4}{c}{\textbf{Metrics (\%)}} \\
\hline
\textbf{$w_{app}$} & \textbf{$w_{geo}$} & \textbf{HOTA $\uparrow$} & \textbf{DetA $\uparrow$} & \textbf{AssA $\uparrow$} & \textbf{LocA $\uparrow$} \\
\hline
1.0 & 0.0 & 51.18 & 30.55 & \textbf{93.12} & 62.18 \\
0.0 & 1.0 & 49.70 & 33.71 & 79.53 & 62.18 \\
\hline
0.7 & 0.3 & 52.13 & 38.24 & 77.14 & 62.18 \\
0.6 & 0.4 & 52.91 & 38.52 & 78.90 & 62.18 \\
\textbf{0.5} & \textbf{0.5} & \textbf{53.13} & 38.39 & 80.16 & 62.18 \\
0.4 & 0.6 & 52.95 & \textbf{38.63} & 78.86 & 62.18 \\
0.3 & 0.7 & 52.83 & 37.84 & 80.22 & 62.18 \\
\hline
\end{tabular}
\end{table}

\subsection{Comparison with State-of-the-Art}
To assess the effectiveness of the proposed calibration-free framework, we compare our results against the top-performing methods from the AI City Challenge 2024 (Track 1). All baseline methods, including the challenge winner Team Yachiyo \cite{10678538_Yoshida}, SJTU-Lenovo \cite{10677892_sjtu_posetrack}, Nota \cite{10678365_nota}, Fraunhofer IOSB \cite{10678409_IOSB}, UW-ETRI \cite{10678220_uw-etri}, and ARV \cite{10678185_arv}, utilized ground-truth camera calibration matrices provided by the dataset as indicated in Table \ref{tab:sota_comparison}.

Operating strictly offline and without any prior geometric knowledge, our calibration-free approach achieved a HOTA of 53.13\%, successfully outperforming calibrated baseline methods like ARV (51.05\%). These results suggest that geometry inferred directly from visual data can serve as a practical alternative to explicit calibration, particularly in scenarios where manual setup is infeasible.

\begin{table*}[t]
\caption{Comparison with state-of-the-art methods on AI City Challenge 2024 (Track 1). All baseline methods utilize ground-truth camera calibration matrices. The proposed method is calibration-free.}
\label{tab:sota_comparison}
\centering
\begin{tabular}{l c c c}
\hline
\textbf{Method} & \textbf{HOTA $\uparrow$} & \textbf{Online} & \textbf{Primary Strength} \\
\hline
Team Yachiyo \cite{10678538_Yoshida} & \textbf{71.94} & No  & High-accuracy clustering and overlap suppression. \\
SJTU-Lenovo \cite{10677892_sjtu_posetrack} & 67.22 & Yes & Geometric consistency and Re-ID correction. \\
Nota \cite{10678365_nota} & 60.93 & Yes & Robust 3D spatio-temporal association. \\
Fraunhofer IOSB \cite{10678409_IOSB} & 60.88 & Yes & Unified 3D world-coordinate tracking. \\
UW-ETRI \cite{10678220_uw-etri} & 57.14 & Yes & Global-local feature fusion for identity matching. \\
ARV \cite{10678185_arv} & 51.05 & Yes & Graph partitioning within sliding temporal window. \\
\hline
\textbf{Ours (Proposed)} & 53.13 & No & 3D reconstruction and pose-guided 3D lifting. \\
\hline
\end{tabular}
\end{table*}

\section{Conclusion}
We presented a calibration-free 3D multi-camera tracking framework that eliminates the need for explicit camera calibration by leveraging transformer-based geometric reconstruction. By integrating pose-guided 3D lifting with joint appearance–geometry clustering, the proposed system maintains spatial coherence and identity consistency across views in complex indoor environments.

Experiments on the AI City Challenge 2024 benchmark demonstrate that competitive 3D tracking performance can be achieved without access to ground-truth camera parameters. While the method does not yet match the highest-performing fully calibrated systems, it establishes a strong baseline for scalable and deployment-friendly 3D tracking.

Several limitations remain. Absolute scale ambiguity may arise without external references, and depth estimation can degrade in texture-poor regions. Future work will explore adaptive scale recovery mechanisms and temporal smoothing strategies to improve stability. We believe calibration-free approaches represent a promising direction toward large-scale, real-world 3D multi-camera analytics.

\bibliographystyle{plainnat}
\bibliography{bibliography}

\begin{thebibliography}{31}
\providecommand{\natexlab}[1]{#1}
\providecommand{\url}[1]{\texttt{#1}}
\expandafter\ifx\csname urlstyle\endcsname\relax
  \providecommand{\doi}[1]{doi: #1}\else
  \providecommand{\doi}{doi: \begingroup \urlstyle{rm}\Url}\fi

\bibitem[Aharon et~al.(2022)Aharon, Orfaig, and
  Bobrovsky]{aharon2022botsortrobustassociationsmultipedestrian}
Nir Aharon, Roy Orfaig, and Ben-Zion Bobrovsky.
\newblock Bot-sort: Robust associations multi-pedestrian tracking.
\newblock \emph{arXiv preprint arXiv:2206.14651}, 2022.

\bibitem[Bewley et~al.(2016)Bewley, Ge, Ott, Ramos, and Upcroft]{Bewley_2016}
Alex Bewley, Zongyuan Ge, Lionel Ott, Fabio Ramos, and Ben Upcroft.
\newblock Simple online and realtime tracking.
\newblock In \emph{2016 IEEE International Conference on Image Processing
  (ICIP)}, page 3464–3468. IEEE, September 2016.
\newblock \doi{10.1109/icip.2016.7533003}.
\newblock URL \url{http://dx.doi.org/10.1109/ICIP.2016.7533003}.

\bibitem[Cherdchusakulchai et~al.(2024)Cherdchusakulchai, Phimsiri,
  Trairattanapa, Tungjitnob, Kudisthalert, Kiawjak, Thamwiwatthana,
  Borisuitsawat, Tosawadi, Choppradit, Mahakijdechachai, Vatathanavaro, Saetan,
  and Suttichaya]{10678185_arv}
Riu Cherdchusakulchai, Sasin Phimsiri, Visarut Trairattanapa, Suchat
  Tungjitnob, Wasu Kudisthalert, Pornprom Kiawjak, Ek~Thamwiwatthana, Phawat
  Borisuitsawat, Teepakorn Tosawadi, Pakcheera Choppradit, Kasisdis
  Mahakijdechachai, Supawit Vatathanavaro, Worawit Saetan, and Vasin
  Suttichaya.
\newblock Online multi-camera people tracking with spatial-temporal mechanism
  and anchor-feature hierarchical clustering.
\newblock In \emph{2024 IEEE/CVF Conference on Computer Vision and Pattern
  Recognition Workshops (CVPRW)}, pages 7198--7207, 2024.
\newblock \doi{10.1109/CVPRW63382.2024.00715}.

\bibitem[Ciaparrone et~al.(2020)Ciaparrone, Luque~Sánchez, Tabik, Troiano,
  Tagliaferri, and Herrera]{Ciaparrone_2020_survey}
Gioele Ciaparrone, Francisco Luque~Sánchez, Siham Tabik, Luigi Troiano,
  Roberto Tagliaferri, and Francisco Herrera.
\newblock Deep learning in video multi-object tracking: A survey.
\newblock \emph{Neurocomputing}, 381:\penalty0 61–88, March 2020.
\newblock ISSN 0925-2312.
\newblock \doi{10.1016/j.neucom.2019.11.023}.
\newblock URL \url{http://dx.doi.org/10.1016/j.neucom.2019.11.023}.

\bibitem[Farhadi et~al.(2018)Farhadi, Redmon,
  et~al.]{redmon2018yolov3incrementalimprovement}
Ali Farhadi, Joseph Redmon, et~al.
\newblock Yolov3: An incremental improvement.
\newblock In \emph{Computer vision and pattern recognition}, volume 1804, pages
  1--6. Springer Berlin/Heidelberg, Germany, 2018.

\bibitem[Ge et~al.(2021)Ge, Liu, Wang, Li, and
  Sun]{ge2021yoloxexceedingyoloseries}
Zheng Ge, Songtao Liu, Feng Wang, Zeming Li, and Jian Sun.
\newblock Yolox: Exceeding yolo series in 2021.
\newblock \emph{arXiv preprint arXiv:2107.08430}, 2021.

\bibitem[Hartley and Zisserman(2004)]{Hartley_Zisserman_2004}
Richard Hartley and Andrew Zisserman.
\newblock \emph{Multiple View Geometry in Computer Vision}.
\newblock Cambridge University Press, 2 edition, 2004.

\bibitem[He et~al.(2016)He, Zhang, Ren, and
  Sun]{he2015deepresiduallearningimage}
Kaiming He, Xiangyu Zhang, Shaoqing Ren, and Jian Sun.
\newblock Deep residual learning for image recognition.
\newblock In \emph{Proceedings of the IEEE conference on computer vision and
  pattern recognition}, pages 770--778, 2016.

\bibitem[Huang et~al.(2023)Huang, Yang, Jiang, Kim, Lee, Kim, Ramkumar,
  Mullapudi, Jang, Huang, et~al.]{huang2023enhancingmulticamerapeopletracking}
Hsiang-Wei Huang, Cheng-Yen Yang, Zhongyu Jiang, Pyong-Kun Kim, Kyoungoh Lee,
  Kwangju Kim, Samartha Ramkumar, Chaitanya Mullapudi, In-Su Jang, Chung-I
  Huang, et~al.
\newblock Enhancing multi-camera people tracking with anchor-guided clustering
  and spatio-temporal consistency id re-assignment.
\newblock In \emph{Proceedings of the IEEE/CVF Conference on Computer Vision
  and Pattern Recognition}, pages 5239--5249, 2023.

\bibitem[Khanam and Hussain(2024)]{khanam2024yolov5deeplookinternal}
Rahima Khanam and Muhammad Hussain.
\newblock What is yolov5: A deep look into the internal features of the popular
  object detector.
\newblock \emph{arXiv preprint arXiv:2407.20892}, 2024.

\bibitem[Kim et~al.(2024)Kim, Shin, Park, and Choi]{10678365_nota}
Jeongho Kim, Wooksu Shin, Hancheol Park, and Donghyuk Choi.
\newblock Cluster self-refinement for enhanced online multi-camera people
  tracking.
\newblock In \emph{2024 IEEE/CVF Conference on Computer Vision and Pattern
  Recognition Workshops (CVPRW)}, pages 7190--7197, 2024.
\newblock \doi{10.1109/CVPRW63382.2024.00714}.

\bibitem[Li et~al.(2019)Li, Wang, Zhu, Mao, Fang, and
  Lu]{Li_2019_CVPR_crowdpose}
Jiefeng Li, Can Wang, Hao Zhu, Yihuan Mao, Hao-Shu Fang, and Cewu Lu.
\newblock Crowdpose: Efficient crowded scenes pose estimation and a new
  benchmark.
\newblock In \emph{Proceedings of the IEEE/CVF Conference on Computer Vision
  and Pattern Recognition (CVPR)}, June 2019.

\bibitem[Lin et~al.(2014)Lin, Maire, Belongie, Hays, Perona, Ramanan,
  Doll{\'a}r, and Zitnick]{lin2014microsoft_coco}
Tsung-Yi Lin, Michael Maire, Serge Belongie, James Hays, Pietro Perona, Deva
  Ramanan, Piotr Doll{\'a}r, and C~Lawrence Zitnick.
\newblock Microsoft coco: Common objects in context.
\newblock In \emph{European conference on computer vision}, pages 740--755.
  Springer, 2014.

\bibitem[Luiten et~al.(2020)Luiten, Osep, Dendorfer, Torr, Geiger,
  Leal-Taix\'e, and Leibe]{Luiten_2020_hota}
Jonathon Luiten, Aljosa Osep, Patrick Dendorfer, Philip Torr, Andreas Geiger,
  Laura Leal-Taix\'e, and Bastian Leibe.
\newblock Hota: A higher order metric for evaluating multi-object tracking.
\newblock \emph{International Journal of Computer Vision}, 129\penalty0
  (2):\penalty0 548–578, October 2020.
\newblock ISSN 1573-1405.
\newblock \doi{10.1007/s11263-020-01375-2}.
\newblock URL \url{http://dx.doi.org/10.1007/s11263-020-01375-2}.

\bibitem[{MMPose Contributors}(2020)]{mmpose2020}
{MMPose Contributors}.
\newblock Openmmlab pose estimation toolbox and benchmark.
\newblock \url{https://github.com/open-mmlab/mmpose}, 2020.

\bibitem[Ren et~al.(2015)Ren, He, Girshick, and
  Sun]{ren2016fasterrcnnrealtimeobject}
Shaoqing Ren, Kaiming He, Ross Girshick, and Jian Sun.
\newblock Faster r-cnn: Towards real-time object detection with region proposal
  networks.
\newblock \emph{Advances in neural information processing systems}, 28, 2015.

\bibitem[Ristani et~al.(2016)Ristani, Solera, Zou, Cucchiara, and
  Tomasi]{ristani2016performancemeasuresdataset}
Ergys Ristani, Francesco Solera, Roger Zou, Rita Cucchiara, and Carlo Tomasi.
\newblock Performance measures and a data set for multi-target, multi-camera
  tracking.
\newblock In \emph{European conference on computer vision}, pages 17--35.
  Springer, 2016.

\bibitem[Schönberger and Frahm(2016)]{7780814}
Johannes~L. Schönberger and Jan-Michael Frahm.
\newblock Structure-from-motion revisited.
\newblock In \emph{2016 IEEE Conference on Computer Vision and Pattern
  Recognition (CVPR)}, pages 4104--4113, 2016.
\newblock \doi{10.1109/CVPR.2016.445}.

\bibitem[Specker(2024)]{10678409_IOSB}
Andreas Specker.
\newblock Ocmctrack: Online multi-target multi-camera tracking with corrective
  matching cascade.
\newblock In \emph{2024 IEEE/CVF Conference on Computer Vision and Pattern
  Recognition Workshops (CVPRW)}, pages 7236--7244, 2024.
\newblock \doi{10.1109/CVPRW63382.2024.00719}.

\bibitem[Sun et~al.(2019)Sun, Xiao, Liu, and Wang]{sun2019deep_hrnet}
Ke~Sun, Bin Xiao, Dong Liu, and Jingdong Wang.
\newblock Deep high-resolution representation learning for human pose
  estimation.
\newblock In \emph{Proceedings of the IEEE conference on computer vision and
  pattern recognition}, pages 5693--5703, 2019.

\bibitem[Tian et~al.(2019)Tian, Shen, Chen, and
  He]{tian2019fcosfullyconvolutionalonestage}
Zhi Tian, Chunhua Shen, Hao Chen, and Tong He.
\newblock Fcos: Fully convolutional one-stage object detection.
\newblock In \emph{Proceedings of the IEEE/CVF international conference on
  computer vision}, pages 9627--9636, 2019.

\bibitem[Wang et~al.(2025)Wang, Chen, Karaev, Vedaldi, Rupprecht, and
  Novotny]{wang2025vggtvisualgeometrygrounded}
Jianyuan Wang, Minghao Chen, Nikita Karaev, Andrea Vedaldi, Christian
  Rupprecht, and David Novotny.
\newblock Vggt: Visual geometry grounded transformer.
\newblock In \emph{Proceedings of the Computer Vision and Pattern Recognition
  Conference}, pages 5294--5306, 2025.

\bibitem[Wang et~al.(2024{\natexlab{a}})Wang, Anastasiu, Tang, Chang, Yao,
  Zheng, Rahman, Arya, Sharma, Chakraborty, et~al.]{wang20248thaicitychallenge}
Shuo Wang, David~C Anastasiu, Zheng Tang, Ming-Ching Chang, Yue Yao, Liang
  Zheng, Mohammed~Shaiqur Rahman, Meenakshi~S Arya, Anuj Sharma, Pranamesh
  Chakraborty, et~al.
\newblock The 8th ai city challenge.
\newblock In \emph{Proceedings of the IEEE/CVF Conference on Computer Vision
  and Pattern Recognition}, pages 7261--7272, 2024{\natexlab{a}}.

\bibitem[Wang et~al.(2024{\natexlab{b}})Wang, Leroy, Cabon, Chidlovskii, and
  Revaud]{wang2024dust3rgeometric3dvision}
Shuzhe Wang, Vincent Leroy, Yohann Cabon, Boris Chidlovskii, and Jerome Revaud.
\newblock Dust3r: Geometric 3d vision made easy.
\newblock In \emph{Proceedings of the IEEE/CVF conference on computer vision
  and pattern recognition}, pages 20697--20709, 2024{\natexlab{b}}.

\bibitem[Wojke et~al.(2017)Wojke, Bewley, and
  Paulus]{wojke2017simpleonlinerealtimetracking}
Nicolai Wojke, Alex Bewley, and Dietrich Paulus.
\newblock Simple online and realtime tracking with a deep association metric.
\newblock In \emph{2017 IEEE international conference on image processing
  (ICIP)}, pages 3645--3649. IEEE, 2017.

\bibitem[Xie et~al.(2024)Xie, Ni, Yang, Zhang, Chen, Zhang, and
  Ma]{10677892_sjtu_posetrack}
Zhenyu Xie, Zelin Ni, Wenjie Yang, Yuang Zhang, Yihang Chen, Yang Zhang, and
  Xiao Ma.
\newblock A robust online multi-camera people tracking system with geometric
  consistency and state-aware re-id correction.
\newblock In \emph{2024 IEEE/CVF Conference on Computer Vision and Pattern
  Recognition Workshops (CVPRW)}, pages 7007--7016, 2024.
\newblock \doi{10.1109/CVPRW63382.2024.00694}.

\bibitem[Yang et~al.(2024)Yang, Huang, Kim, Jiang, Kim, Huang, Du, and
  Hwang]{10678220_uw-etri}
Cheng-Yen Yang, Hsiang-Wei Huang, Pyong-Kun Kim, Zhongyu Jiang, Kwang-Ju Kim,
  Chung-I Huang, Haiqing Du, and Jenq-Neng Hwang.
\newblock An online approach and evaluation method for tracking people across
  cameras in extremely long video sequence.
\newblock In \emph{2024 IEEE/CVF Conference on Computer Vision and Pattern
  Recognition Workshops (CVPRW)}, pages 7037--7045, 2024.
\newblock \doi{10.1109/CVPRW63382.2024.00697}.

\bibitem[Yoshida et~al.(2024)Yoshida, Okubo, Fujii, Amakata, and
  Yamashita]{10678538_Yoshida}
Ryuto Yoshida, Junichi Okubo, Junichiro Fujii, Masazumi Amakata, and Takayoshi
  Yamashita.
\newblock Overlap suppression clustering for offline multi-camera people
  tracking.
\newblock In \emph{2024 IEEE/CVF Conference on Computer Vision and Pattern
  Recognition Workshops (CVPRW)}, pages 7153--7162, 2024.
\newblock \doi{10.1109/CVPRW63382.2024.00710}.

\bibitem[Zhang(2000)]{888718}
Z.~Zhang.
\newblock A flexible new technique for camera calibration.
\newblock \emph{IEEE Transactions on Pattern Analysis and Machine
  Intelligence}, 22\penalty0 (11):\penalty0 1330--1334, 2000.
\newblock \doi{10.1109/34.888718}.

\bibitem[Zhou et~al.(2019{\natexlab{a}})Zhou, Yang, Cavallaro, and
  Xiang]{zhou2019omniscalefeaturelearningperson}
Kaiyang Zhou, Yongxin Yang, Andrea Cavallaro, and Tao Xiang.
\newblock Omni-scale feature learning for person re-identification.
\newblock In \emph{Proceedings of the IEEE/CVF international conference on
  computer vision}, pages 3702--3712, 2019{\natexlab{a}}.

\bibitem[Zhou et~al.(2019{\natexlab{b}})Zhou, Wang, and
  Kr{\"a}henb{\"u}hl]{zhou2019objectspoints}
Xingyi Zhou, Dequan Wang, and Philipp Kr{\"a}henb{\"u}hl.
\newblock Objects as points.
\newblock \emph{arXiv preprint arXiv:1904.07850}, 2019{\natexlab{b}}.

\end{thebibliography}

\end{document}